# Existence and Finiteness Conditions for Risk-Sensitive Planning: Results and Conjectures

**Yaxin Liu**
Department of Computer Sciences
University of Texas at Austin
Austin, TX 78712-0233
yxliu@cs.utexas.edu

**Sven Koenig**
Computer Science Department
University of Southern California
Los Angeles, CA 90089-0781
skoenig@usc.edu

## Abstract

Decision-theoretic planning with risk-sensitive planning objectives is important for building autonomous agents or decision-support systems for real-world applications. However, this line of research has been largely ignored in the artificial intelligence and operations research communities since planning with risk-sensitive planning objectives is more complicated than planning with risk-neutral planning objectives. To remedy this situation, we derive conditions that guarantee that the optimal expected utilities of the total plan-execution reward exist and are finite for fully observable Markov decision process models with non-linear utility functions. In case of Markov decision process models with both positive and negative rewards, most of our results hold for stationary policies only, but we conjecture that they can be generalized to non-stationary policies.

## 1  Introduction

Decision-theoretic planning is important since real-world applications need to cope with uncertainty. Many decision-theoretic planners use fully observable Markov decision process models (MDPs) from operations research [11] to represent probabilistic planning problems. However, most of them minimize the expected total plan-execution cost or, synonymously, maximize the expected total reward (MER planners). This risk-neutral planning objective and similar simplistic planning objectives often do not take the preferences of human decision makers sufficiently into account, for example, their risk attitudes in planning domains with huge wins or losses of money, equipment, or human life. This means that they are not ideally suited for real-world planning, including space applications [14], environmental applications [2] and business applications [6]. In this paper, we therefore provide a first step toward a comprehensive foundation of risk-sensitive planning by deriving conditions that guarantee that the optimal expected utilities of the total reward exist and are finite for fully observable Markov decision process models with non-linear utility functions.

Table 1: An Example of Risk Sensitivity

|  | Probability | Reward | Expected Reward | Utility | Expected Utility |
|---|---|---|---|---|---|
| Choice 1 | 50% | $10,000,000 | $5,000,000 | $-0.050$ | $-0.525$ |
|  | 50% | $ 0 |  | $-1.000$ |  |
| Choice 2 | 100% | $ 4,500,000 | $4,500,000 | $-0.260$ | $-0.260$ |

## 2  Risk Attitudes and Utility Theory

Human decision makers are typically risk-sensitive and thus do not maximize the expected total reward in planning domains with huge wins or losses. Table 1 shows an example for which many human decision makers prefer Choice 2 over Choice 1 even though its expected total reward is lower. They are risk-averse and thus accept a reduction in expected total reward for a reduction in variance. Utility theory [12] suggests that this behavior is rational because human decision makers maximize the expected utility of the total reward. Utility functions map total rewards to the corresponding finite utilities and are monotonically non-decreasing in the total reward. They capture the risk attitudes of human decision makers [10]. Linear utility functions result in maximizing the expected total reward and characterize risk-neutral human decision makers (MER planning objective), while non-linear utility functions characterize risk-sensitive human decision makers (MEU planning objective). In particular, concave utility functions characterize risk-averse human decision makers ("insurance holders"), and convex utility functions characterize risk-seeking human decision makers ("lottery players"). For example, if a risk-averse human decision maker has the concave exponential utility function $U(w) = -0.9999997^w$ and thus associates the utilities shown in Table 1 with the total rewards of the two choices, then Choice 2 maximizes their expected utility and should thus be chosen by them. On the other hand, MER planners pick Choice 1. The human decision maker would thus be extremely unhappy with them with 50 percent probability, which motivates our desire to build MEU planners.

## 3 Markov Decision Process Models

We study MEU planners that use MDPs to represent probabilistic planning problems. Formally, an MDP is a 4-tuple $(S, A, P, r)$ of a state space $S$, an action space $A$, a set of transition probabilities $P$ and a set of finite (immediate) rewards $r$. If an agent executes $a \in A$ in $s \in S$, then it incurs reward $r(s, a, s')$ and transitions to $s' \in S$ with probability $P(s'|s, a)$. An MDP is called finite if its state space and action space are both finite. We assume throughout this paper that the MDPs are finite since decision-theoretic planners typically use finite MDPs.

The kinds of MDPs that decision-theoretic planners typically use tend to have goal states that need to be reached [3]. The MDP in Figure 1(a) gives an example. Its transitions are labeled with their rewards followed by their probabilities. The rewards of the two actions in $s^1$ are negative because they correspond to costs. $s^2$ is the goal state, in which only one action can be executed and its execution incurs zero reward and leaves the state unchanged. To achieve generality, however, we do not make any assumptions about the structure of the MDPs or their rewards. For example, we do not make any assumptions about how the structure of the MDPs and their rewards encode the goal states or about whether the goal states can be reached. Neither do we make any assumptions about whether all of the rewards are positive, negative or zero. We avoid such assumptions because MDPs can mix positive rewards (which model, for example, rewards for reaching goal states) and negative rewards (which model, for example, costs for executing actions).

## 4 Planning Horizons and Policies

The number of time steps that a planner plans for is called its (planning) horizon. A history at time step $t$ is a sequence $h_t = (s_0, a_0, \cdots, s_{t-1}, a_{t-1}, s_t)$ of states and actions from the initial state $s_0$ to the current state $s_t$. The set of all histories at time step $t$ is $H_t = (S \times A)^t \times S$. A trajectory is an element of $H_\infty$ for infinite horizons and $H_T$ for finite horizons, where we use $T \geq 1$ to denote the last time step of the finite horizon.

Decision-theoretic planners determine a decision rule for every time step within the horizon. A decision rule determines which action the agent should execute in its current state. A randomized history-dependent (HR) decision rule, the most general decision rule, at time step $t$ is a mapping $d_t : H_t \to \mathcal{P}(A)$, where $\mathcal{P}(A)$ is the set of probability distributions over $A$. A Markovian decision rule is a history-dependent decision rule whose actions depend only on the current state. A randomized Markovian decision rule at time step $t$ is a mapping $d_t : S \to \mathcal{P}(A)$. A deterministic Markovian decision rule at time step $t$ is a mapping $d_t : S \to A$.

A policy $\pi$ is a sequence of decision rules $d_t$, one for every time step $t$ within the horizon. We use $\Pi^{\text{HR}}$ to denote the set of all policies with HR decision rules, which is the same as the set of all policies. A policy is called stationary if $d_t = d$ for all time steps $t$ and $d$ is a Markovian decision rule. We use $\Pi^{\text{SR}}$ to denote the set of all stationary randomized (SR) policies, which is the same as the set of all stationary policies, and $\Pi^{\text{SD}}$ to denote the set of all stationary deterministic (SD) policies. It holds that $\Pi^{\text{SD}} \subseteq \Pi^{\text{SR}} \subseteq \Pi^{\text{HR}}$.

## 5 Optimal Values and Optimal Policies

For probabilistic planning problems with a finite horizon $T$, the expected utility of the total reward obtained by starting in $s \in S$ and following $\pi \in \Pi$ is defined to be
$$v_{U,T}^\pi(s) = E^{s,\pi}\left[U\left(\sum_{t=0}^{T-1} r_t\right)\right],$$
where $r_t = r(s_t, a_t, s_{t+1})$ and the expectation $E^{s,\pi}$ is taken over all possible trajectories. The expected utilities exist and are bounded because the number of trajectories is finite for finite MDPs. MEU planners then need to determine the maximal expected utilities of the total reward $v_{U,T}^*(s) = \sup_{\pi \in \Pi} v_{U,T}^\pi(s)$ and a policy that achieves them. The maximal expected utilities exist and are finite because the expected utilities are bounded.

For probabilistic planning problems with an infinite horizon, the expected utility of the total reward obtained by starting in $s \in S$ and following $\pi \in \Pi$ is defined to be
$$v_U^\pi(s) = \lim_{T \to \infty} v_{U,T}^\pi(s) = \lim_{T \to \infty} E^{s,\pi}\left[U\left(\sum_{t=0}^{T-1} r_t\right)\right]. \quad (1)$$

The expected utilities exist iff the limit exists, that is, the limit is a finite number, positive infinity or negative infinity. MEU planners then need to determine the maximal expected utilities of the total reward $v_U^*(s) = \sup_{\pi \in \Pi} v_U^\pi(s)$ and a policy that achieves them. To simplify our terminology, we refer to the expected utilities $v_U^\pi(s)$ as the values for $\pi \in \Pi$ and to the maximal expected utilities $v_U^*(s)$ as the optimal values. A policy $\pi \in \Pi$ is optimal iff $v_U^\pi(s) = v_U^*(s)$ for all $s \in S$.

## 6 Existence and Finiteness Conditions

In this paper, we derive conditions that guarantee that the optimal values exist and are finite for MDPs with non-linear utility functions, as a first step toward a comprehensive foundation of risk-sensitive planning. When we say that the values exist (or are finite), we mean that they have this property for all policies and all states. Similarly, when we say that the optimal values exist (or are finite), we mean that they have this property for all states.

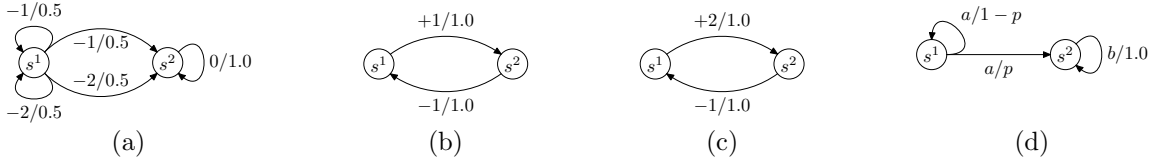

Figure 1: Example MDPs

It is important that the optimal values exist since MEU planners determine a policy that achieves them. There are cases where the optimal values do not exist, as the MDP in Figure 1(b) illustrates. An agent that starts in $s^1$ receives the following sequence of rewards for its only policy: $+1, -1, +1, -1, \ldots$, and consequently the following sequence of total rewards: $+1, 0, +1, 0, \ldots$, which oscillates. Thus, the limit in Eq. (1) does not exist for any utility function with $U(0) \neq U(1)$, and the optimal value of $s^1$ does not exist either. A similar argument holds for $s^2$ as well.

It is also important that the optimal values be finite. There are cases where the optimal values are not finite, as the MDP in Figure 1(a) illustrates. The MDP has two stationary deterministic policies. $\pi_1$ assigns the top action to $s^1$, and $\pi_2$ assigns the bottom action to $s^1$. Consider $U(w) = -\left(\frac{1}{2}\right)^w$. The values are

$$v_U^{\pi_1}(s^1) = \sum_{t=1}^{\infty}\left[-\left(\tfrac{1}{2}\right)^{(-1)t} \cdot (1/2)^t\right] = -\sum_{t=1}^{\infty} 1 = -\infty,$$

$$v_U^{\pi_2}(s^1) = \sum_{t=1}^{\infty}\left[-\left(\tfrac{1}{2}\right)^{(-2)t} \cdot (1/2)^t\right] = -\sum_{t=1}^{\infty} 2^t = -\infty,$$

and $v_U^{\pi_1}(s^2) = v_U^{\pi_2}(s^2) = -1$. Thus, the optimal values are $v_U^*(s^1) = \max(-\infty, -\infty) = -\infty$ and $v_U^*(s^2) = \max(-1, -1) = -1$. All trajectories have identical probabilities for both policies, but the total reward and thus also the utility of each trajectory is larger for policy $\pi_1$ than policy $\pi_2$. Thus, policy $\pi_1$ should be preferred over policy $\pi_2$ for all utility functions. Policy $\pi_2$ thus demonstrates that a policy that achieves the optimal values and thus is optimal according to our definition is not always the best one. The problem is that policies with infinite values are indistinguishable, and thus the optimal values need to be finite to compare policies in a meaningful way. This example also shows that the optimal values are not guaranteed to be finite even if all policies reach a goal state with probability one. Furthermore, the optimal values of both states are, for example, finite for $U(w) = w$ and thus any policy that achieves them is indeed the best one for this utility function, which shows that the problem can exist for some utility functions but not others.

The values exist and are bounded if one uses discounting, that is, assumes that a reward obtained at some time step is worth only a fraction of the same reward obtained one time step earlier. Discounting is a way of modeling interest on investments. Such interest often does not exist, for example, for human life. This is fortunate because discounting makes it very difficult to find optimal policies for non-linear utility functions [13]. For example, it is known that all optimal policies can be non-stationary (and thus difficult to find) for positive or negative MDPs with exponential utility functions if discounting is used [7]. On the other hand, there always exists an optimal stationary deterministic policy for positive and negative MDPs with exponential utility functions if discounting is not used [1, 4]. In the following, we therefore do not use discounting.

## 7 Existing Results

We first review conditions that guarantee that the optimal values exist and are finite. These conditions have been obtained for MDPs with linear and exponential utility functions. We then use these results to identify similar conditions for more general MDPs and more general utility functions.

### 7.1 Linear Utility Functions

We first consider linear utility functions. They are of the form $U(w) = w$ and characterize risk-neutral human decision makers. We omit the subscript $U$ for linear utility functions.

#### 7.1.1 Positive MDPs

MDPs for which C1 holds are called positive [11]. The values $v^\pi(s)$ exist for positive MDPs since $v_T^\pi(s)$ is monotone in $T$. Thus, the optimal values exist as well. The optimal values are finite if C2 holds [11], which bounds the optimal values from above. The optimal values are finite even if $\Pi$ is replaced with $\Pi^{\text{SD}}$ in C2 since there always exists an optimal stationary deterministic policy for linear utility functions [11].

**C1:** For all $s, s' \in S$ and all $a \in A$, $r(s, a, s') \geq 0$.

**C2:** For all $\pi \in \Pi$ and all $s \in S$, $v^\pi(s)$ is finite.

### 7.1.2 Negative MDPs

MDPs for which C3 holds are called negative [11]. Similar to positive MDPs, the values exist for negative MDPs and thus the optimal values exist as well. The optimal values are finite if C4 holds [11], which bounds the optimal values from below. The optimal values are finite even if $\Pi$ is replaced with $\Pi^{SD}$ in C4 since there always exists an optimal stationary deterministic policy for linear utility functions [11].

**C3:** For all $s, s' \in S$ and all $a \in A$, $r(s, a, s') \leq 0$.

**C4:** There exists $\pi \in \Pi$ such that, for all $s \in S$, $v^\pi(s)$ is finite.

### 7.1.3 General MDPs

In general, MDPs can have both positive and negative (as well as zero) rewards. We define the positive part of a real number $r$ to be $r^+ = \max(r, 0)$ and its negative part to be $r^- = \min(r, 0)$. We then obtain the positive part of an MDP by replacing every reward of the MDP with its positive part. We use $v^{+\pi}(s)$ to denote the values of the positive part of an MDP for policy $\pi \in \Pi$. We define the negative part of an MDP and $v^{-\pi}(s)$ in an analogous way.

The values exist and $v^\pi(s) = v^{+\pi}(s) + v^{-\pi}(s)$ for all $s \in S$ and all $\pi \in \Pi$ if C5 holds [11]. Thus, the optimal values exist as well but they are not guaranteed to be finite [11].

**C5:** For all $\pi \in \Pi$ and all $s \in S$, at least one of $v^{+\pi}(s)$ and $v^{-\pi}(s)$ is finite.

The optimal values are finite if C6 and C7 hold [11], which bound the optimal values from above and below, respectively. The optimal values are finite even if $\Pi$ is replaced with $\Pi^{SD}$ in C6 and C7 since there always exists an optimal stationary deterministic policy for linear utility functions [11].

**C6:** For all $\pi \in \Pi$ and all $s \in S$, $v^{+\pi}(s)$ is finite.

**C7:** There exists $\pi \in \Pi$ such that, for all $s \in S$, $v^{-\pi}(s)$ is finite.

Consider $U(w) = w$. The MDP in Figure 1(b) then violates C5. The values do not exist for its only policy $\pi$, as we have argued earlier. It is easy to see that $v^{+\pi}(s^1) = +\infty$ and $v^{-\pi}(s^1) = -\infty$, which violates C5 and illustrates that C5 indeed rules out MDPs whose values do not exist for all policies. The MDP in Figure 1(c) is another MDP that violates C5. The values, however, exist for its only policy $\pi'$. For example, an agent that starts in $s^1$ receives the following sequence of rewards for its only policy: $+2, -1, +2, -1, \ldots$, and consequently the following sequence of total rewards: $+2, +1, +3, +2, +4, +3, \ldots$, which converges toward positive infinity. Thus, the limit in Eq. (1) exists for $\pi'$, that is, the value of $s^1$ exists for $\pi'$. However, it is easy to see that $v^{+\pi'}(s^1) = +\infty$ and $v^{-\pi'}(s^1) = -\infty$, which violates C5 and demonstrates that C5 is not a necessary condition for the values to exist.

## 7.2 Exponential Utility Functions

We now consider exponential utility functions, the most widely used non-linear utility functions [5]. They are of the form $U_e(w) = \iota \gamma^w$ for $\gamma > 0$, where $\iota = \text{sign} \ln \gamma$. If $\gamma > 1$, then the exponential utility function is convex and characterizes risk-seeking human decision makers. If $0 < \gamma < 1$, then the utility function is concave and characterizes risk-averse human decision makers. We use the subscript e instead of $U$ for exponential utility functions.

### 7.2.1 Positive MDPs

The values $v_e^\pi(s)$ exist for positive MDPs since $v_{e,T}^\pi(s)$ is monotone in $T$. Thus, the optimal values exist as well. The optimal values are finite if either $0 < \gamma < 1$, or if $\gamma > 1$ and C8 holds [4]. The optimal values are finite even if $\Pi$ is replaced with $\Pi^{SD}$ since there always exists an optimal stationary deterministic policy for exponential utility functions [4].

**C8:** For all $\pi \in \Pi$ and all $s \in S$, $v_e^\pi(s)$ is finite.

### 7.2.2 Negative MDPs

Similar to positive MDPs, the values exist for negative MDPs and thus the optimal values exist as well. The optimal values are finite if either $\gamma > 1$, or if $0 < \gamma < 1$ and C9 holds [1]. The optimal values are finite even if $\Pi$ is replaced with $\Pi^{SD}$ in C9 since there always exists an optimal stationary deterministic policy for exponential utility functions [1].

**C9:** There exists $\pi \in \Pi$ such that, for all $s \in S$, $v_e^\pi(s)$ is finite.

## 8 New Results: Overview

We now propose several new conditions that guarantee that the optimal values exist and are finite, to cover additional situations. Each condition consists of constraints on the utility function and the MDP. For example, we study utility functions that grow with different rates, namely utility functions that are bounded, linearly bounded, exponentially bounded, and exponential in addition to general utility functions without any constraints other than them being monotonically non-decreasing. We study MDPs whose values satisfy different conditions, including C5 as the weakest constraint, to MDPs whose total rewards are bounded as the strongest constraint.

The following lemmata about stationary policies are key to proving our main results. Because of the space limit, we state our results and provide only intuitive

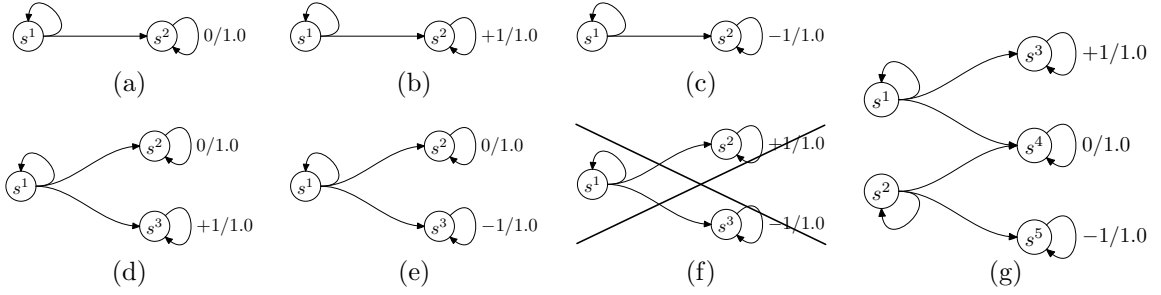

Figure 2: Example MDPs that Illustrate Lemma 1

explanations instead of formal proofs. All proofs are available in [9].

Lemma 1 states implications of C5 for stationary policies. Its proof is a direct application of C5. Stationary policies determine Markov chains. A state of a Markov chain and thus also a state of an MDP for a stationary policy is called recurrent iff the expected number of time steps between visiting the state is finite, otherwise it is called transient. A recurrent class is a maximal set of states that are recurrent and reachable from each other. We use $R^\pi$ to denote the set of recurrent states and $R_i^\pi \subseteq R^\pi$ to denote a recurrent class for a stationary policy $\pi$.

**Lemma 1.** *Assume that C5 holds. Let $\pi \in \Pi^{SR}$. Then, for all $s \in S$ and all recurrent classes $R_i^\pi$,*

**a.** *if $v^\pi(s)$ is finite, then, for all $s' \in S$ that are reachable from $s$ if one follows $\pi$, $v^\pi(s')$ is finite. If $s' \in R_i^\pi$, then the rewards of all transitions within $R_i^\pi$ are zero and, for all $s'' \in R_i^\pi$, $v^\pi(s'') = 0$,*

**b.** *if $v^\pi(s) = \infty$, then, for all $s' \in S$ that are reachable from $s$ if one follows $\pi$, $v^\pi(s') = \infty$ or $v^\pi(s')$ is finite. If $s' \in R_i^\pi$, then either (1) the rewards of all transitions within $R_i^\pi$ are zero and, for all $s'' \in R_i^\pi$, $v^\pi(s'') = 0$ or (2) the rewards of all transitions within $R_i^\pi$ are nonnegative and at least one is positive and, for all $s'' \in R_i^\pi$, $v^\pi(s'') = \infty$, and*

**c.** *if $v^\pi(s) = -\infty$, then, for all $s' \in S$ that are reachable from $s \in S$ if one follows $\pi$, $v^\pi(s') = -\infty$ or $v^\pi(s')$ is finite. If $s' \in R_i^\pi$, then either (1) the rewards of all transitions within $R_i^\pi$ are zero and, for all $s'' \in R_i^\pi$, $v^\pi(s'') = 0$ or (2) the rewards of all transitions within $R_i^\pi$ are nonpositive and at least one is negative and, for all $s'' \in R_i^\pi$, $v^\pi(s'') = -\infty$.*

Lemma 1 classifies the recurrent classes for stationary policies under C5 into three types (zero, positive and negative), depending on whether all states in them have value zero, positive infinity or negative infinity. The MDPs in Figure 2(a)-(c) have one recurrent class of each type. The MDPs in Figure 2(d)-(e) illustrate that it is possible to reach more than one recurrent class from the same state. However, it is impossible that some of them are positive and others are negative. The MDP in Figure 2(f) is therefore impossible. The MDP in Figure 2(g) illustrates that MDPs can have both positive and negative recurrent classes but they cannot be reached from the same state.

Lemma 2 concerns the well-known geometric rate of state evolution [8] for stationary policies, which is important since the rewards accumulate only at a linear rate. The expressions $\rho^t$ suggest correctly that exponential utility functions are often key in the proofs of our main results.

**Lemma 2.** *Let $\pi \in \Pi^{SR}$. Then, for all $s \in S$, there exists $0 < \rho < 1$ such that*

**a.** *there exists $a > 0$ such that, for all $t \geq 0$, $P^{s,\pi}(s_t \notin R^\pi) \leq a\rho^t$,*

**b.** *there exists $b > 0$ such that, for all $t \geq 0$, $P^{s,\pi}(s_t \notin R^\pi, s_{t+1} \in R^\pi) \leq b\rho^t$, and*

**c.** *for all recurrent classes $R_i^\pi$, there exists $c > 0$ such that, for all $t \geq 0$, $P^{s,\pi}(s_t \notin R^\pi, s_{t+1} \in R_i^\pi) \leq c\rho^t$,*

*where $P^{s,\pi}$ is a shorthand for a probability if $s_0 = s$ and one follows $\pi$.*

The MDP in Figure 1(d) illustrates Lemma 2. $s^2$ is the only recurrent state for its only policy $\pi$ if $p > 0$. For all $t \geq 0$, $P^{s^1,\pi}(s_t \neq s^2) = (1-p)^t$, illustrating Lemma 2a, and $P^{s^1,\pi}(s_t \neq s^2, s_{t+1} = s^2) = p(1-p)^t$, illustrating Lemma 2b and Lemma 2c.

We are now ready to give an overview of the proofs of our main results. We start by deriving conditions that guarantee that the values exist for all stationary policies. We use $w_t = \sum_{i=0}^{t-1} r_i$ to denote the total reward up to time step $t$ and $\tau(T)$ with $0 \leq \tau(T) \leq T$ to denote the first time step within horizon $T$ of being in any recurrent state, implying that $\tau(T) = 0$ if $s_0 \in R^\pi$, and $s_{\tau(T)-1} \notin R^\pi$ and $s_{\tau(T)} \in R^\pi$ otherwise. We

can then decompose the values as follows:

$$v_U^\pi(s) = \lim_{T\to\infty} v_{U,T}^\pi(s)$$
$$= \lim_{T\to\infty} E^{s,\pi}\left[U\left(\sum_{t=0}^{T-1} r_t\right)\right] = \lim_{T\to\infty} E^{s,\pi}[U(w_T)]$$
$$= \lim_{T\to\infty} E^{s,\pi}[U(w_T)|s_T \notin R^\pi] \cdot P^{s,\pi}(s_T \notin R^\pi)$$
$$+ \lim_{T\to\infty} E^{s,\pi}[U(w_T)|s_T \in R^\pi] \cdot P^{s,\pi}(s_T \in R^\pi)$$
$$= \lim_{T\to\infty} E^{s,\pi}[U(w_T)|s_T \notin R^\pi] \cdot P^{s,\pi}(s_T \notin R^\pi)$$
$$+ \lim_{T\to\infty} E^{s,\pi}[U(w_{\tau(T)})|s_T \in R^\pi] \cdot P^{s,\pi}(s_T \in R^\pi) \quad (2)$$
$$+ \lim_{T\to\infty} E^{s,\pi}[U(w_T) - U(w_{\tau(T)})|s_T \in R^\pi] \cdot P^{s,\pi}(s_T \in R^\pi).$$

The first limit of Eq. (2) is the contribution of those trajectories that do not enter any recurrent state within horizon $T$, while the second and third limit are the contributions of those trajectories that enter a recurrent state. The second limit is the contribution until any recurrent state is entered for the first time, and the third limit is the contribution after a recurrent state has been entered for the first time. Therefore, the values exist if all three limits exist.

The following lemma proves that the third limit exists for stationary policies under C5. Its proof uses the types of recurrent classes from Lemma 1.

**Lemma 3.** *Assume that C5 holds. For all $\pi \in \Pi^{SR}$ and all $s \in S$,*

$$\lim_{T\to\infty} E^{s,\pi}[U(w_T) - U(w_{\tau(T)})|s_T \in R^\pi] \cdot P^{s,\pi}(s_T \in R^\pi)$$

*exists. Moreover, if $v^\pi(s)$ is finite, the limit is zero; if $v^\pi(s) = +\infty$, the limit is finite if the utility function is bounded from above, and positive infinity otherwise; and if $v^\pi(s) = -\infty$, the limit is finite if the utility function is bounded from below, and negative infinity otherwise.*

Not surprisingly, C5 is part of many of the following theorems in this paper, which establish additional conditions under which the first and second limits of Eq. (2) and thus also the values exist. Their proofs make use of the following properties: The first limit is zero iff the second limit is finite; the limit superior (lim sup) of the first term is positive iff the second limit is positive infinity; and the limit inferior (lim inf) of the first term is negative iff the second limit is negative infinity. The proofs of these properties use Lemma 2 or stronger conditions. We show later in the context of Lemma 5 and Lemma 6 how to prove special cases of these properties.

If the values $v_U^\pi(s)$ exist then the optimal values $v_U^*(s) = \sup_{\pi \in \Pi} v_U^\pi(s)$ also exist. Similarly, if the values are bounded from above for all policies and all states and are finite for at least one policy and all states then the optimal values are finite. Unfortunately, we are often only able to state conditions that guarantee that the values exist for all stationary policies because the counterparts of Lemma 1 and Lemma 2 for non-stationary policies are not known to hold. If it is unknown whether there always exists an optimal stationary policy or known that there does not always exist an optimal stationary policy, then we cannot conclude from this result that the optimal values exist. This is not a problem for linear utility functions, where it is known that there always exists an optimal stationary deterministic policy. Unfortunately, this property does not hold for non-linear utility functions [13]. This is the reason why we listed in Section 7 existing results the few cases where optimal stationary policies are known to exist. However, in many cases, we are able to state additional conditions that guarantee that the optimal values are finite if they exist. The proofs of these properties use the positive part of the MDP to provide an upper bound on the optimal values and the negative part of the MDP to provide a lower bound.

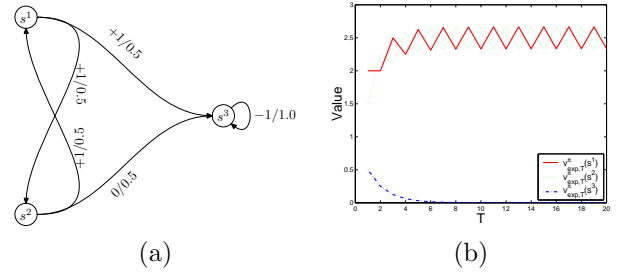

(a) (b)

Figure 3: Example MDP that Illustrates C5

## 9 Exponential Utility Functions

We first discuss exponential utility functions in the context of MDPs with both positive and negative rewards, a situation not yet discussed in Section 7. We use $v_e^{+\pi}(s)$ to denote the values of the positive part of an MDP and $v_e^{+*}(s)$ to denote its optimal values if the utility function is exponential. We define $v_e^{-\pi}(s)$ and $v_e^{-*}(s)$ in an analogous way.

The following lemma shows that the $v_{e,T}^\pi(s)$ can be calculated as matrix powers, where $\gamma$ is the parameter of the exponential utility function and $\iota = \text{sign} \ln \gamma$. $\pi(s,a)$ denotes the probability with which to execute $a \in A$ in $s \in S$ if one follows $\pi \in \Pi^{SR}$. Its proof is by induction.

**Lemma 4.** *Let $\pi \in \Pi^{SR}$ and $D_\pi$ be the matrix whose $(s,s')$-entry is, for all $s, s' \in S$, $D_\pi(s,s') = \sum_{a \in A} \pi(s,a) P(s'|s,a) \gamma^{r(s,a,s')}$. Let $D_\pi^T$ be the $T$-th power of $D_\pi$. Then, for all $s, s' \in S$,*

$$D_\pi^T(s,s') = E^{s,\pi}[\gamma^{w_T}|s_T = s'] \cdot P^{s,\pi}(s_T = s')$$

and $v_{e,T}^\pi(s) = \iota \cdot \sum_{s' \in S} D_\pi^T(s, s')$.

The MDP in Figure 3(a) and its only policy $\pi$ illustrate Lemma 4. For the convex exponential utility function $U(w) = 2^w$, $v_{e,T}^\pi(s)$ can be obtained by calculating the matrix powers of

$$D_\pi = \begin{pmatrix} 0 & 1 & 1 \\ 1 & 0 & \frac{1}{2} \\ 0 & 0 & \frac{1}{2} \end{pmatrix}.$$

They are

$$v_{e,T}^\pi(s^1) = \tfrac{5}{2} - \tfrac{1}{6}(-1)^T - \tfrac{4}{3}\left(\tfrac{1}{2}\right)^T,$$
$$v_{e,T}^\pi(s^2) = \tfrac{5}{2} + \tfrac{1}{6}(-1)^T - \tfrac{5}{3}\left(\tfrac{1}{2}\right)^T,$$
$$v_{e,T}^\pi(s^3) = \left(\tfrac{1}{2}\right)^T,$$

as shown in Figure 3(b). Therefore, $v_e^\pi(s^1)$ and $v_e^\pi(s^2)$ do not exist. For the concave exponential utility function $U(w) = -\left(\tfrac{1}{2}\right)^w$, the same matrix $D_\pi$ results if all rewards of the MDP are negated, and $v_e^\pi(s^1)$ and $v_e^\pi(s^2)$ do not exist either. However, the MDP satisfies C5. Thus, C5 is thus too weak to guarantee that the values exist.

The conditions that can eliminate such MDPs are different for convex and concave exponential utility functions. Before considering them separately, we present some results about stationary policies that hold for both of them and are key for our later results by establishing that the first and second limits of Eq. (2) exist. The following lemmata relate them.

**Lemma 5.** Let $\pi \in \Pi^{SR}$. For all $s \notin R^\pi$,

$$\lim_{T \to \infty} E^{s,\pi}\left[\gamma^{w_T} \mid s_T \notin R^\pi\right] \cdot P^{s,\pi}(s_T \notin R^\pi) = 0$$

iff $\lim_{T \to \infty} E^{s,\pi}\left[\gamma^{w_{\tau(T)}} \mid s_T \in R^\pi\right] \cdot P^{s,\pi}(s_T \in R^\pi)$ is finite.

**Lemma 6.** Let $\pi \in \Pi^{SR}$. For all $s \notin R^\pi$,

$$\limsup_{T \to \infty} E^{s,\pi}\left[\gamma^{w_T} \mid s_T \notin R^\pi\right] \cdot P^{s,\pi}(s_T \notin R^\pi) > 0,$$

iff $\lim_{T \to \infty} E^{s,\pi}\left[\gamma^{w_{\tau(T)}} \mid s_T \in R^\pi\right] \cdot P^{s,\pi}(s_T \in R^\pi) = +\infty$.

The proofs of Lemma 5 and Lemma 6 are based on the following lemma, whose proof is by induction.

**Lemma 7.** Let $\pi \in \Pi^{SR}$ and $\hat{D}_\pi$ be the matrix whose $(s, s')$-entry is, for all $s, s' \in S$,

$$\hat{D}_\pi(s, s') = \begin{cases} D_\pi(s, s') & s \notin R^\pi \\ 1 & s = s', s \in R^\pi \\ 0 & s \neq s', s \in R^\pi. \end{cases}$$

Let $\hat{D}_\pi^T$ be the $T$-th power of $\hat{D}_\pi$. Then, for all $s, s' \in S$,

$$\hat{D}_\pi^T(s, s') = \begin{cases} E^{s,\pi}\left[\gamma^{w_T} \mid s_T = s'\right] \cdot P^{s,\pi}(s_T = s') \\ \qquad s, s' \notin R^\pi \\ E^{s,\pi}\left[\gamma^{w_{\tau(T)}} \mid s_{\tau(T)} = s'\right] \cdot P^{s,\pi}(s_{\tau(T)} = s') \\ \qquad s \notin R^\pi, s' \in R^\pi \\ 1 \qquad s = s', s \in R^\pi \\ 0 \qquad s \neq s', s \in R^\pi. \end{cases}$$

We can order the indices of $\hat{D}_\pi$ so that $\hat{D}_\pi = \begin{pmatrix} A_\pi & B_\pi \\ \mathbf{0} & \mathbf{1} \end{pmatrix}$ for some matrices $A_\pi$ and $B_\pi$, where $\mathbf{1}$ is an identity matrix and $\mathbf{0}$ is a zero matrix. The top rows then correspond to the transient states, and the bottom rows correspond to the recurrent states. Then,

$$\hat{D}_\pi^T = \begin{pmatrix} A_\pi^T & \left(\sum_{t=0}^{T-1} A_\pi^t\right) \cdot B_\pi \\ \mathbf{0} & \mathbf{1} \end{pmatrix},$$

and Lemma 5 and Lemma 6 follow from the fact that the limit of $\hat{D}_\pi^T$ exists as $T \to \infty$ iff $A_\pi^T \to \mathbf{0}$ since the limit of $\left(\sum_{t=0}^{T-1} A_\pi^t\right) \cdot B_\pi$ then exists as well.

### 9.1 Convex Exponential Utility Functions

We first consider the case where the utility function is convex exponential. Then, Theorem 8 states a condition under which the values exist for all stationary policies $\pi$. Its proof makes use of Lemma 5, Lemma 6 and the types of the recurrent classes from Lemma 1.

**C10:** For all $\pi \in \Pi$ and all $s \in S$, at least one of $v_e^{+\pi}(s)$ and $v^{-\pi}(s)$ is finite.

**Theorem 8.** *Assume that the utility function is convex exponential and C10 holds. Then, for all $\pi \in \Pi^{SR}$ and all $s \in S$, $v_e^\pi(s)$ exists.*

Consider again the MDP in Figure 3(a) and its only policy $\pi$. For the convex exponential utility function $U(w) = 2^w$, $v_{e,T}^{+\pi(s)}$ can be obtained by calculating the matrix powers of the positive part of the MDP

$$\begin{pmatrix} 0 & 1 & 1 \\ 1 & 0 & \frac{1}{2} \\ 0 & 0 & 1 \end{pmatrix}$$

according to Lemma 4. They are

$$v_{e,T}^{+\pi}(s^1) = \tfrac{9}{8} - \tfrac{1}{8}(-1)^T + \tfrac{3}{4}T,$$
$$v_{e,T}^{+\pi}(s^2) = \tfrac{7}{8} + \tfrac{1}{8}(-1)^T + \tfrac{3}{4}T,$$
$$v_{e,T}^{+\pi}(s^3) = 1.$$

Therefore, $v_e^{+\pi}(s^1) = v_e^{+\pi}(s^2) = +\infty$. On the other hand, it follows from Lemma 1 that $v^{-\pi}(s^3) = -\infty$, and thus $v^{-\pi}(s^1) = v^{-\pi}(s^2) = -\infty$. Consequently, the MDP violates C10, and C10 is sufficiently strong to eliminate this MDP.

We conjecture that Theorem 8 also holds for all policies. Since it is currently unknown whether there always exists an optimal stationary policy, it is also unknown whether the optimal values exist. Assume that C11 holds, the utility function is convex exponential, and the optimal values exist. Then, the optimal values are finite since C11 bounds them from above. This is the case because the positive MDP satisfies C8 and thus $v_e^{+*}(s)$ is finite. Furthermore, for all $\pi \in \Pi$ and all $s \in S$,

$$v_{e,T}^\pi(s) = E^{s,\pi}\left[U_e\left(\sum_{t=0}^{T-1} r_t\right)\right]$$
$$\leq E^{s,\pi}\left[U_e\left(\sum_{t=0}^{T-1} r_t^+\right)\right] = v_{e,T}^{+\pi}(s).$$

Taking the limit as $T \to \infty$ shows that $v_e^\pi(s) \leq v_e^{+\pi}(s)$. Therefore, $v_e^*(s) \leq v_e^{+*}(s) < +\infty$.

**C11:** For all $\pi \in \Pi$ and all $s \in S$, $v_e^{+\pi}(s)$ is finite.

### 9.2 Concave Exponential Utility Functions

We now consider the case where the utility function is concave exponential. The results and proofs for concave exponential utility functions are analogous to the ones for convex exponential utility functions.

**C12:** For all $\pi \in \Pi$ and all $s \in S$, at least one of $v^{+\pi}(s)$ and $v_e^{-\pi}(s)$ is finite.

**Theorem 9.** *Assume that the utility function is concave exponential and C12 holds. For all $\pi \in \Pi^{SR}$ and all $s \in S$, $v_e^\pi(s)$ exists.*

We conjecture that Theorem 9 also holds for all policies. Since it is currently unknown whether there always exists an optimal stationary policy, it is also unknown whether the optimal values exist. Assume that the utility function is concave exponential, C13 hold, and the optimal values exist. Then, the optimal values are finite since C13 bounds them from below. This is the case because for the $\pi$ from C13 and all $s \in S$,

$$v_{e,T}^\pi(s) = E^{s,\pi}\left[U_e\left(\sum_{t=0}^{T-1} r_t\right)\right]$$
$$\geq E^{s,\pi}\left[U_e\left(\sum_{t=0}^{T-1} r_t^-\right)\right] = v_{e,T}^{-\pi}(s).$$

Taking the limit as $T \to \infty$ shows that $v_e^*(s) \geq v_e^\pi(s) \geq v_e^{-\pi}(s) > -\infty$.

**C13:** There exists $\pi \in \Pi$ such that, for all $s \in S$, $v_e^{-\pi}(s)$ is finite.

## 10 General Utility Functions

We now consider non-linear utility functions that are more general than exponential utility functions. Such utility functions are, for example, necessary to model risk attitudes that change with the total reward.

### 10.1 Positive and Negative MDPs

We first consider positive and negative MDPs. The values $v_U^\pi(s)$ exist since $v_{U,T}^\pi(s)$ is monotone in $T$. Thus, the optimal values exist as well. Theorem 10 and Theorem 11 state conditions under which the values and optimal values are finite for positive MDPs, and Theorem 12 and Theorem 13 state conditions under which they are finite for negative MDPs. The theorems hold since $v_U^\pi(s)$ and $v_U^*(s)$ are dominated by $v^\pi(s)$ and $v^*(s)$ (for Theorem 10 and Theorem 12) or by $v_e^\pi(s)$ and $v_e^*(s)$ (for Theorem 11 and Theorem 13), respectively, where $\gamma$ is the parameter of the exponential utility function.

**Theorem 10.** *Assume that C1 and C2 hold and there exist $C, D > 0$ such that $U(w) \leq Cw + D$ for all $w \geq 0$. Then, for all $\pi \in \Pi$ and all $s \in S$, $v_U^\pi(s)$ and $v_U^*(s)$ exist and are finite.*

**Theorem 11.** *Assume that C1 and C8 hold for some $\gamma > 1$ and there exist $C, D > 0$ such that $U(w) \leq C\gamma^w + D$ for this $\gamma$ and all $w \geq 0$. Then, for all $\pi \in \Pi$ and all $s \in S$, $v_U^\pi(s)$ and $v_U^*(s)$ exist and are finite.*

**Theorem 12.** *Assume that C3 and C4 hold for some $\pi \in \Pi$ and there exist $C, D > 0$ such that $U(w) \geq -Cw - D$ for all $w \leq 0$. Then, for this $\pi$ and all $s \in S$, $v_U^\pi(s)$ and $v_U^*(s)$ exist and are finite.*

**Theorem 13.** *Assume that C3 and C9 hold for some $\pi \in \Pi$ and some $\gamma$ with $0 < \gamma < 1$ and there exist $C, D > 0$ such that $U(w) \geq -C\gamma^w - D$ for this $\gamma$ and all $w \leq 0$. Then, for this $\pi$ and all $s \in S$, $v_U^\pi(s)$ and $v_U^*(s)$ are finite.*

### 10.2 General MDPs

We now consider MDPs with both positive and negative rewards.

#### 10.2.1 Bounded Functions

We first consider the case where the utility function is bounded, that is, there exist finite $U^+$ and $U^-$ such that $U^- \leq U(w) \leq U^+$ for all $w$. Then, $v_{U,T}^\pi(s)$ is bounded as $T \to \infty$ but can oscillate, in which case the values do not exist. Theorem 14 states a condition under which the values exist for all stationary policies $\pi$. Its proof shows that the first limit of Eq. (2) is zero and the second limit exists, making use of Lemma 2 and $U^+$ and $U^-$ as bounds.

**Theorem 14.** *Assume that the utility function is bounded and C5 holds. Then, for all $\pi \in \Pi^{SR}$ and all $s \in S$, $v_U^\pi(s)$ exists and is finite.*

We conjecture that Theorem 14 also holds for all policies. Since it is known that there does not always exist an optimal stationary policy, it is currently unknown whether the optimal values exist. Assume that the utility function is bounded and the optimal values exist. Then, the optimal values are finite since the values are bounded.

### 10.2.2 Linearly Bounded Functions

We now consider the case where the utility function is linearly bounded, that is, there exist $C, D > 0$ such that $U(w) \leq Cw + D$ for all $w \geq 0$ and $U(w) \geq -Cw - D$ for all $w \leq 0$. Then, Theorem 15 states a condition under which the values exist for all stationary policies. Its proof shows that the first limit of Eq. (2) is zero and the second limit exists, making use of Lemma 2 and the linear functions as bounds.

**Theorem 15.** *Assume that the utility function is linearly bounded and C5 holds. Then, for all $\pi \in \Pi^{SR}$ and all $s \in S$, $v_U^\pi(s)$ exists.*

We conjecture that Theorem 15 also holds for all policies. Since it is known that there does not always exist an optimal stationary policy, it is currently unknown whether the optimal values exist. Assume that the utility function is linearly bounded, C6 and C7 hold, and the optimal values exist. Then, the optimal values are finite since C6 bounds them from above and C7 bounds them from below. The proof is similar to the one for exponential utility functions.

### 10.2.3 Exponentially Bounded Functions

We now consider the case where the utility function is exponentially bounded, that is, there exist $C, D > 0$, $\gamma_+ > 1$ and $0 < \gamma_- < 1$ such that $U(w) \leq C\gamma_+^w + D$ for all $w \geq 0$ and $U(w) \geq -C\gamma_-^w - D$ for all $w \leq 0$. Then, Theorem 16 states a condition under which the values exist for all stationary policies $\pi$, where $\text{e}(\gamma_+)$ refers to the exponential utility function with parameter $\gamma_+$ and $\text{e}(\gamma_-)$ refers to the exponential utility function with parameter $\gamma_-$. Its proof shows that the first limit of Eq. (2) is zero and the second limit exists, making use of Lemma 6 and the exponential functions as bounds.

**C14:** For all $\pi \in \Pi$ and all $s \in S$, $v_{\text{e}(\gamma_+)}^{+\pi}(s)$ and $v_{\text{e}(\gamma_-)}^{-\pi}(s)$ are finite.

**Theorem 16.** *Assume that the utility function is exponentially bounded and C14 holds. Then, for all $\pi \in \Pi^{SR}$ and all $s \in S$, $v_U^\pi(s)$ exists and is finite.*

However, C14 is very restrictive. It excludes, for example, MDPs with cycles with acyclic optimal policies. An analogy to the case of linearly bounded utility functions suggests that one might be able to use the weaker condition C15.

**C15:** For all $\pi \in \Pi$ and all $s \in S$, at least one of $v_{\text{e}(\gamma_+)}^{+\pi}(s)$ and $v_{\text{e}(\gamma_-)}^{-\pi}(s)$ is finite.

Assume that the utility function is exponentially bounded and C15 holds. Then, we conjecture that the values exist but currently cannot even prove this conjecture for all stationary policies. Since it is known that there does not always exist an optimal stationary policy, it is currently unknown whether the optimal values exist. Assume that the utility function is exponentially bounded, C11 and C13 hold for $\gamma_+$ and $\gamma_-$, respectively, and the optimal values exist. Then, the optimal values are finite since C11 bounds them from above and C13 bounds them from below. The proof is similar to the one for exponential utility functions.

### 10.2.4 Bounded Total Rewards

Finally, we consider the case where the total reward is bounded from above or below. We use $H_T^{s,\pi}$ to denote the set of trajectories with finite horizon $T$ obtained by starting in $s \in S$ and following $\pi \in \Pi$. We define $w(h_T) = \sum_{t=0}^{T-1} r_t$ for all $h_T = (s_0, a_0, \cdots, s_{T-1}, a_{T-1}, s_T) \in H_T^{s,\pi}$. We also define $v_{\max,T}^\pi(s) = \max_{h_T \in H_T^{s,\pi}} w(h_T)$, $v_{\max}^\pi(s) = \lim_{T \to \infty} v_{\max,T}^\pi(s)$, $v_{\min,T}^\pi(s) = \min_{h_T \in H_T^{s,\pi}} w(h_T)$, and $v_{\min}^\pi(s) = \lim_{T \to \infty} v_{\min,T}^\pi(s)$. Then, $v_{\max}^{+\pi}(s)$ and $v_{\min}^{-\pi}(s)$ exist since $v_{\max,T}^\pi(s)$ and $v_{\min,T}^\pi(s)$ are monotone in $T$. Theorem 17 states a condition under which the values exist.

**C16:** For all $\pi \in \Pi$ and all $s \in S$, at least one of $v_{\max}^{+\pi}(s)$ and $v_{\min}^{-\pi}(s)$ is finite.

**Theorem 17.** *Assume that C16 holds. Then, for all $\pi \in \Pi$ and all $s \in S$, $v_U^\pi(s)$ exists. Thus, $v_U^*(s)$ exists as well.*

Assume that C17 and C18 hold. Then, the optimal values are finite since C17 bounds them from above and C18 bounds them from below. The proof is similar to the one for exponential utility functions. C17 and C18 are, for example, satisfied for acyclic MDPs where plan execution is guaranteed to end in absorbing states but are satisfied for some MDPs with cycles as well.

**C17:** For all $\pi \in \Pi$ and all $s \in S$, $v_{\max}^{+\pi}(s)$ is finite.

**C18:** There exists $\pi \in \Pi$ such that, for all $s \in S$, $v_{\min}^{-\pi}(s)$ is finite.

## 11 Conclusions

In this paper, we derived conditions that guarantee that the optimal expected utilities exist and are finite for MDPs with non-linear utility functions. We derived results for positive MDPs, negative MDPs and general MDPs with both positive and negative rewards. Table 2 summarizes our existence results for general MDPs and various kinds of utility functions. A checkmark represents that the expected utilities exist and the optimal expected utilities therefore exist as well. A checkmark with a question mark represents that the expected utilities are known to exist for all stationary policies but it is currently unknown whether they exist for all policies and thus also whether the optimal expected utilities exist. A question mark represents that it is currently unknown whether the expected utilities exist even for all stationary policies. A cross represents that the optimal expected utilities are known not to exist in all cases. A dash represents that the result for this case should follow from other cases since C16 implies C15, which in turn implies C5. In future work, we intend to remove the question marks in the table. More importantly, the results in the paper on the existence and finiteness of the optimal expected utilities provide only a first step toward a comprehensive foundation of risk-sensitive planning. In future work, we also intend to study the structure of optimal policies and basic computational procedures for obtaining them.

Table 2: Summary of Results for General MDPs

| Kind of Utility Functions | C16 | C15 | C5 |
|---|---|---|---|
| Bounded | — | — | ✓? |
| Linearly Bounded | — | — | ✓? |
| Exponentially Bounded | — | ? | ✗ |
| General | ✓ | (✗) | ✗ |

**Acknowledgments**

We thank Craig Tovey and Anton Kleywegt, two operations researchers, for lots of advice and mentoring while the first author was writing his dissertation and their careful proofreading of the proofs. Without their expertise, it would have been impossible to write this paper. This research was partly supported by NSF awards to Sven Koenig under contracts IIS-9984827 and IIS-0098807 and an IBM fellowship to Yaxin Liu. The views and conclusions contained in this document are those of the authors and should not be interpreted as representing the official policies, either expressed or implied, of the sponsoring organizations, agencies, companies or the U.S. government.

# References

[1] M. G. Ávila-Godoy. *Controlled Markov Chains with Exponential Risk-Sensitive Criteria: Modularity, Structured Policies and Applications*. PhD thesis, Department of Mathematics, University of Arizona, 1999.

[2] J. Blythe. *Planning under Uncertainty in Dynamic Domains*. PhD thesis, School of Computer Science, Carnegie Mellon University, 1997.

[3] C. Boutilier, T. Dean, and S. Hanks. Decision-theoretic planning: Structural assumptions and computational leverage. *Journal of Artificial Intelligence Research*, 11:1–94, 1999.

[4] R. Cavazos-Cadena and R. Montes-de-Oca. Nearly optimal policies in risk-sensitive positive dynamic programming on discrete spaces. *Mathematics Methods of Operations Research*, 52:133–167, 2000.

[5] J. L. Corner and P. D. Corner. Characteristics of decisions in decision analysis practice. *The Journal of the Operational Research Society*, 46:304–314, 1995.

[6] R. T. Goodwin, R. Akkiraju, and F. Wu. A decision-support system for quote-generation. In *Proceedings of the Fourteenth Conference on Innovative Applications of Artificial Intelligence (IAAI-02)*, pages 830–837, 2002.

[7] S. C. Jaquette. A utility criterion for Markov decision processes. *Management Science*, 23(1):43–49, 1976.

[8] J. G. Kemeny and J. L. Snell. *Finite Markov Chains*. D. Van Nostrand Company, 1960.

[9] Y. Liu. *Decision Theoretic Planning under Risk-Sensitive Planning Objectives*. PhD thesis, Georgia Institute of Technology, 2005. Available as http://www.cc.gatech.edu/~yxliu/thesis.pdf.

[10] J. W. Pratt. Risk aversion in the small and in the large. *Econometrica*, 32(1-2):122–136, 1964.

[11] M. L. Puterman. *Markov Decision Processes: Discrete Stochastic Dynamic Programming*. John Wiley & Sons, 1994.

[12] J. von Neumann and O. Morgenstern. *Theory of Games and Economic Behavior*. Princeton University Press, 1944.

[13] D. J. White. Utility, probabilistic constraints, mean and variance of discounted rewards in Markov decision processes. *OR Spektrum*, 9:13–22, 1987.

[14] S. Zilberstein, R. Washington, D. S. Bernstein, and A.-I. Mouaddib. Decision-theoretic control of planetary rovers. In M. Beetz, J. Hertzberg, M. Ghallab, and M. E. Pollack, editors, *Advances in Plan-Based Control of Robotic Agents*, volume 2466 of *Lecture Notes in Computer Science*, pages 270–289. Springer, 2002.